\documentclass[letterpaper, 10 pt, conference]{ieeeconf}  

\IEEEoverridecommandlockouts                              

\usepackage{amssymb}  
\usepackage{amsmath} 
\usepackage{url}
\usepackage{wrapfig}
\usepackage{graphicx}
\usepackage{subcaption}

\usepackage{tikz}

\usepackage{siunitx}
\usepackage{xspace}
\usepackage{tabularx}
\usepackage{booktabs}
\usepackage{cite}
\usepackage{balance}
\usepackage{arydshln}  
\usepackage[bookmarks=true,colorlinks=true,citecolor=blue,linkcolor=blue,urlcolor=blue]{hyperref}
\usepackage{float}
\usepackage[dvipsnames]{xcolor}
\definecolor{steelblue}{HTML}{4682B4}
\usepackage{dblfloatfix}
\usepackage{multirow,graphicx,array}

\newif\ifshowcomments
\showcommentstrue
\showcommentsfalse 
\ifshowcomments
    \usepackage[textsize=scriptsize]{todonotes}
    \newcommand{\fix}[1]{{\color{red} #1}}
\else
    \usepackage[disable,textsize=scriptsize]{todonotes}
    \newcommand{\fix}[1]{}
\fi

\newcommand{\todomw}[1]{\todo[fancyline,color=green!40]{MW: #1}\xspace}

\newcommand{\todotj}[1]{\todo[fancyline,color=blue!40]{TJ: #1}\xspace}
\newcommand{\todointj}[1]{\todo[inline,color=blue!40]{TJ: #1}}

\newcommand{\todojd}[1]{\todo[fancyline,color=orange!40]{JD: #1}\xspace}

\newcommand{\g}[1]{\textcolor{ForestGreen}{#1}}

\newcommand{\plucker}[0]{Pl{\"u}cker\xspace}

\makeatletter
\let\NAT@parse\undefined
\makeatother
\usepackage[square, numbers, sort]{natbib}

\title{\LARGE \bf
Do You Know Where Your Camera Is? \\
View-Invariant Policy Learning with Camera Conditioning
}

\author{%
Tianchong Jiang$^{1}$,
Jingtian Ji$^{1}$,
Xiangshan Tan$^{1}$,
Jiading Fang$^{2,\ast}$, \\
Anand Bhattad$^3$,
Vitor Guizilini$^{4,\dagger}$,
Matthew R.~Walter$^{1,\dagger}$
\thanks{$^{1}$Toyota Technological Institute at Chicago (TTIC), Chicago, IL, USA. {\tt\small tianchongj@ttic.edu}}%
\thanks{$^{2}$Waymo, USA. $^{\ast}$Work completed while at TTIC.}%
\thanks{$^{3}$Johns Hopkins University.}%
\thanks{$^{4}$Toyota Research Institute (TRI), USA.}%
\thanks{$^{\dagger}$ Equal advising}
}

\begin{document}

\maketitle
\thispagestyle{empty}
\pagestyle{empty}

\begin{abstract}

We study view-invariant imitation learning by explicitly conditioning policies on camera extrinsics.
Using \plucker embeddings of per-pixel rays, we show that conditioning on extrinsics significantly improves generalization across viewpoints for standard behavior cloning policies, including ACT, Diffusion Policy, and SmolVLA.
To evaluate policy robustness under realistic viewpoint shifts, we introduce six manipulation tasks in RoboSuite and ManiSkill that pair ``fixed'' and ``randomized'' scene variants, decoupling background cues from camera pose.
Our analysis reveals that policies without extrinsics often infer camera pose using visual cues from static backgrounds in fixed scenes. This shortcut collapses when workspace geometry or camera placement shifts.
Conditioning on extrinsics restores performance and yields robust RGB-only control without depth. We release the tasks, demonstrations, and code to facilitate reproducibility and future research. 
Code and project materials are available at \href{https://ripl.github.io/know_your_camera/}{ripl.github.io/know\_your\_camera}.


\end{abstract}

\section{Introduction} \label{sec:introduction}


Significant recent attention has been devoted to imitation learning as a means of enabling robots to perform diverse and complex manipulation tasks~\cite{Octo2023, OpenX2023, OpenVLA2024, PiZero2023}. 
These policies usually directly operate on RGB images and are trained with fixed third-person viewpoints. 
However, when deployed in the real world, it is often not possible to position the cameras to exactly match their training pose due to environmental constraints. Consequently, many such models fail when testing on different viewpoints.
The problem is exacerbated when the robot's embodiment changes, which inherently involves diverse camera configurations and viewpoints, making viewpoint invariance essential for cross-embodiment transfer~\cite{Zakka2022}.

\begin{figure}[!t]
  \centering
  \includegraphics[width=0.95\linewidth]{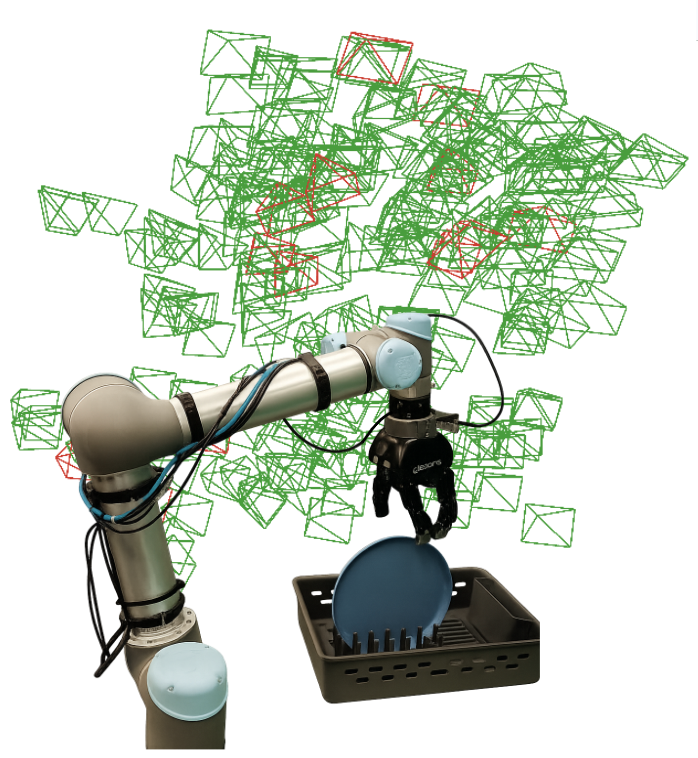}
  \caption{Visualization of camera poses in the real-robot experiment. Training cameras are visualized in green, and test cameras are visualized in red.}
  \label{fig:real_robot}
\end{figure}
Concurrently, accurate camera geometry is increasingly accessible. Such information can be obtained directly from the datasets~\cite{DROID2023} used for training, or estimated through classic methods such as hand-eye calibration~\cite{TsaiLenz1989}, visual-SLAM and structure-from-motion methods~\cite{schoenberger2016sfm, ORB_SLAM2017, HTCViveTracking2018, VINSMono2018, fastmap2025}, or modern learning-based methods that handle sparse or dynamic views~\cite{vggt, huang2025vipe}. Neglecting this readily available geometric information is a significant missed opportunity for improving robot learning.


\todojd{I moved the discussion of camera randomization into related works section, I recognize the importance to discuss this dominant approach. While your counter point of not seeing the robot during randomization is realistic, I believe the fundamental discussion here should be between implicit or explicit camera conditioning. We should offer some potential benefits, and proceed to unroll our study.}

To achieve viewpoint invariance, a policy must be aware of the camera's geometry relative to the robot. This raises a fundamental question: should the end-to-end policy \textit{implicitly} infer camera geometries during training, or should the robot policy model \textit{explicitly} condition on camera geometries when they are readily available?

While the architectural simplicity of end-to-end learning is appealing, especially in data-rich domains like language modeling, robotics is notoriously data-scarce. Expecting a single model to simultaneously\todomw{I reworded this since we said that estimatinc camera pose was a ``difficult, unsolved problem,'' however we just suggested that it is not particularly hard.} 
estimate camera pose while also learning complex manipulation policies poses a serious trade-off between learnability and data efficiency. Therefore, we investigate the benefits of \textit{explicit} camera conditioning. Assuming access to camera geometry, we study its effect on the robustness of robot policies under viewpoint variations.  

More specifically, our contributions are the following:

\begin{itemize}
    \item \textbf{A Novel Camera Conditioning Method.} We propose an approach to effectively condition common imitation learning policies on explicit camera geometries with per-pixel \plucker embeddings.
    \item \textbf{Comprehensive Empirical Analysis.} 
    We show the benefits of explicit camera conditioning for policy generalization to view variations, and identify key factors affecting performance, including action space, random cropping, early vs.\ late conditioning, the effect of the pretrained visual encoder, and the scaling law with respect to the number of camera views.
    \item \textbf{New Benchmarks for Viewpoint Generalization.} We introduce benchmark tasks---three in RoboSuite and three in ManiSkill---targeting challenging viewpoint generalization, and the corresponding demonstrations.
\end{itemize}




\section{Related Work} \label{sec:related-work}



We review imitation learning, methods for view-invariant policies, and approaches to encoding camera geometry.

\subsection{Imitation Learning}

Imitation learning offers a direct and sample-efficient approach to teaching robots complex manipulation skills by learning from expert demonstrations. 
To mitigate compounding errors from single-step policies, \citet{ACT2023} introduced Action Chunking with Transformers (ACT), where policies predict entire action sequences rather than single actions, effectively reducing the task horizon. 
A key challenge is modeling the multi-modal nature of human demonstrations, as a task can often be solved in multiple valid ways. 
To address this multi-modality, methods like ACT, Diffusion Policy (DP)~\cite{DiffusionPolicy2023}, and Behavioral Transformers (BeT)~\cite{Shafiullah2022} learn distributions over actions rather than regress to the mean. 
A recent trend is to scale policies using large, diverse datasets. 
Models like RT-1~\cite{RT12023}, RT-2~\cite{Brohan2023}, Octo~\cite{Octo2023}, OpenVLA~\cite{OpenVLA2024}, Pi-0~\cite{PiZero2023}, LBM~\cite{LBM}, and SmolVLA~\cite{SmolVLA2025} demonstrate that training a single, high-capacity model on large, heterogeneous datasets results in improved generalization to novel variations of seen tasks and policies that can be efficiently fine-tuned for new settings.

\subsection{Learning View-Invariant Robot Policies}

Achieving policy invariance to camera viewpoint is a long-standing goal in robot learning. Existing methods can be broadly categorized by their input modality and whether they learn invariance implicitly or explicitly.

\paragraph{3D and Depth-Based Methods} When depth is available, RGB-D camera feeds can be lifted into explicit 3D representations that are naturally robust to camera pose~\cite{Peri2024PCWM}.
One line of work transforms sensor data into voxel grids or learned feature fields~\cite{peract, peract2, gnfactor, act3d, pdfactor}. Another approach synthesizes canonical views for the policy, decoupling perception from the physical camera placement~\cite{rvt, rvt2}. Others explore inherently view-invariant SE(3)-equivariant architectures~\cite{equibot, equact} or operate directly on point clouds~\cite{dp3, 3d-diffuser-actor}.
\begin{figure}[!t]
  \centering
  \begin{subfigure}{0.5\textwidth}
      \includegraphics[width=\linewidth]{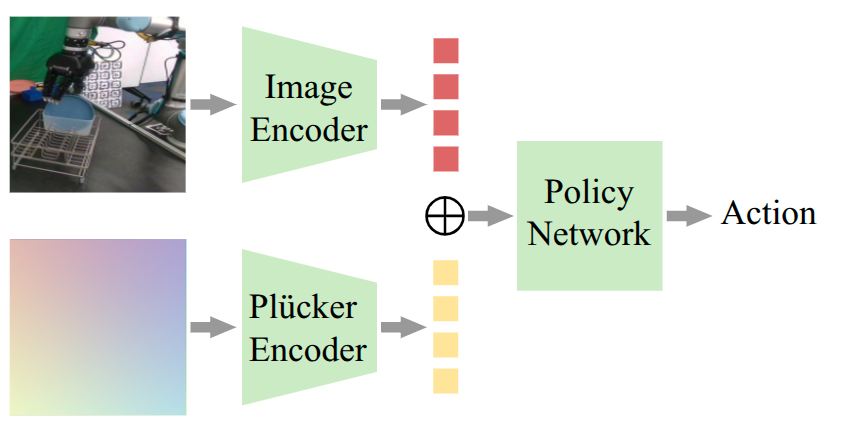}
      \caption{Policy with a pretrained encoder}\label{fig:arch-with-pretrained}
  \end{subfigure}
\vspace{4mm}
  \begin{subfigure}{0.5\textwidth}
    \includegraphics[width=\linewidth]{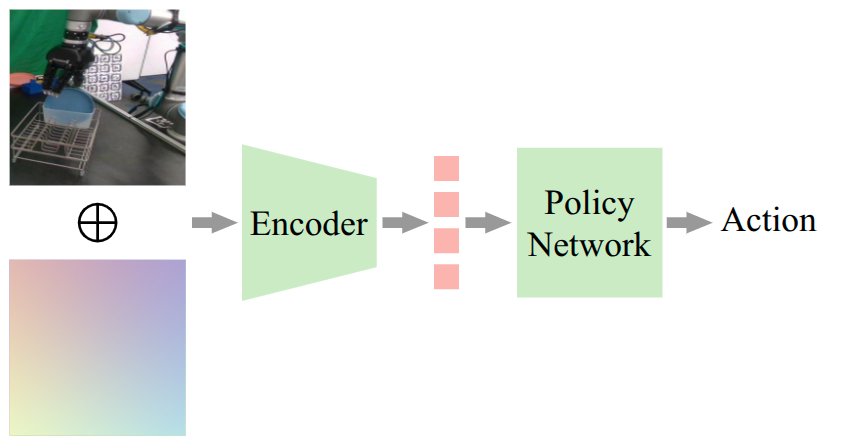}
    \caption{Policy without a pretrained encoder}\label{fig:arch-without-pretrained}
  \end{subfigure}

  \caption{We propose two ways by which to encode \plucker ray-maps for policies \subref{fig:arch-with-pretrained} with and \subref{fig:arch-without-pretrained} without a pretrained encoder. The $\bigoplus$ sign indicates channel-wise concatenation.}
  \label{fig:arch}
\end{figure}

\paragraph{RGB-Only Methods} While effective, the 3D methods above rely on high-quality depth sensors, which remain expensive and less ubiquitous than standard cameras. We specifically focus on RGB-based policies because RGB-only input arguably represents the minimal sensing modality for general-purpose robots. Furthermore, vast quantities of internet data and most large-scale behavior cloning datasets are predominantly RGB-based~\cite{R3M2022, RT12023}.
Within RGB-only methods, a popular strategy for improving robustness is to implicitly learn invariance by randomizing camera viewpoints during data collection or training. This approach has been adopted for several real-world robotics datasets~\cite{DROID2023, RoboSet2023, RH20T2022, RoboNet2019, BridgeDataV22023}. Similarly, synthetic data augmentation techniques re-render scenes from novel camera extrinsics using techniques like novel view synthesis~\cite{VISTA2024, roviaug} or 3D Gaussian Splatting~\cite{robosplat}. However, this strategy raises a fundamental question: if the robot is barely visible under heavy viewpoint randomization, how can a policy implicitly determine the camera's pose relative to its own action frame?

\begin{figure*}[!t]
  \centering
  \setlength{\tabcolsep}{3pt}
  \begin{tabular}{cccc:ccc}
    \raisebox{40pt}{\rotatebox{90}{\textbf{Fixed}}} &
    \begin{subfigure}{0.15\textwidth}
      \includegraphics[width=\linewidth]{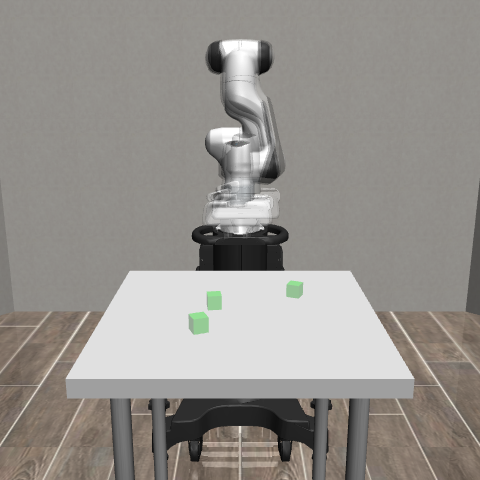}
      \caption*{Lift}
    \end{subfigure} &
    \begin{subfigure}{0.15\textwidth}
      \includegraphics[width=\linewidth]{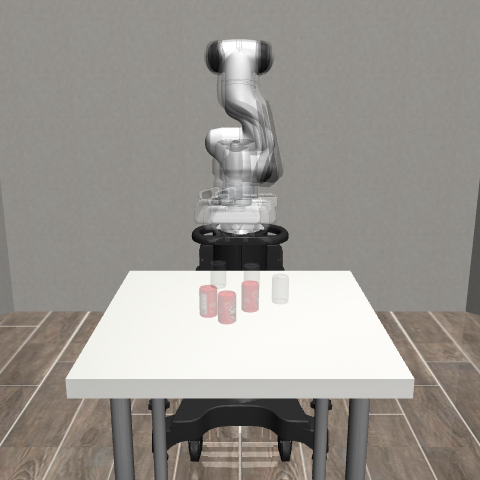}
      \caption*{Pick Place Can}
    \end{subfigure} &
    \begin{subfigure}{0.15\textwidth}
      \includegraphics[width=\linewidth]{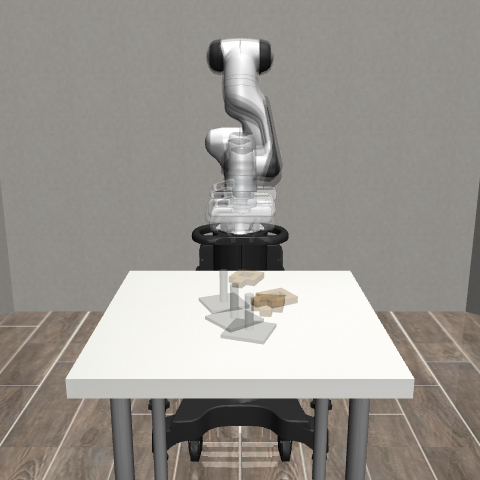}
      \caption*{Assembly Square}
    \end{subfigure} &
    \begin{subfigure}{0.15\textwidth}
      \includegraphics[width=\linewidth]{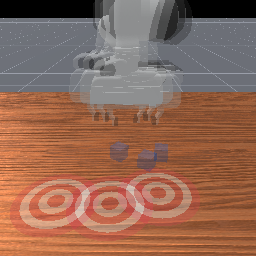}
      \caption*{Push}
    \end{subfigure} &
    \begin{subfigure}{0.15\textwidth}
      \includegraphics[width=\linewidth]{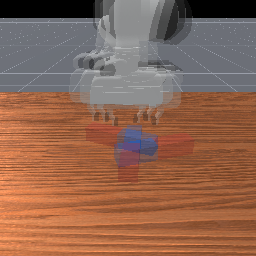}
      \caption*{Lift Upright}
    \end{subfigure} &
    \begin{subfigure}{0.15\textwidth}
      \includegraphics[width=\linewidth]{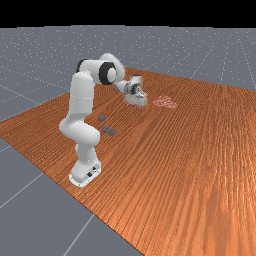}
      \caption*{Roll Ball}
    \end{subfigure}
    \\
    \raisebox{26pt}{\rotatebox{90}{\textbf{Randomized}}} &
    \begin{subfigure}{0.15\textwidth}
      \includegraphics[width=\linewidth]{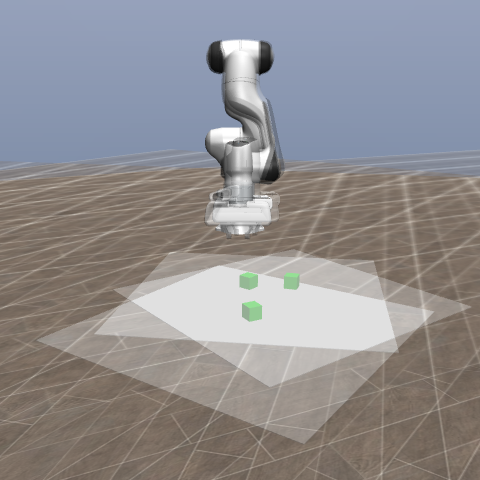}
      \caption*{Lift}
    \end{subfigure} &
    \begin{subfigure}{0.15\textwidth}
      \includegraphics[width=\linewidth]{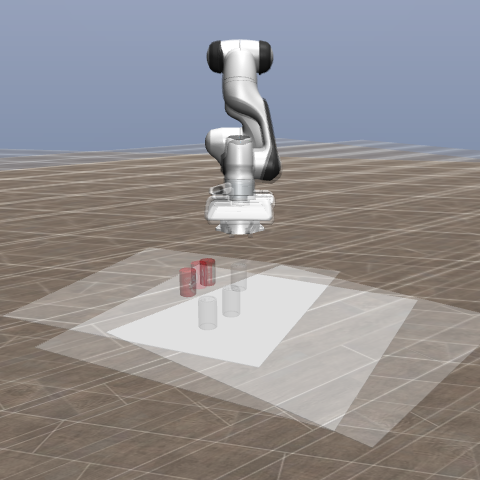}
      \caption*{Pick Place Can}
    \end{subfigure} &
    \begin{subfigure}{0.15\textwidth}
      \includegraphics[width=\linewidth]{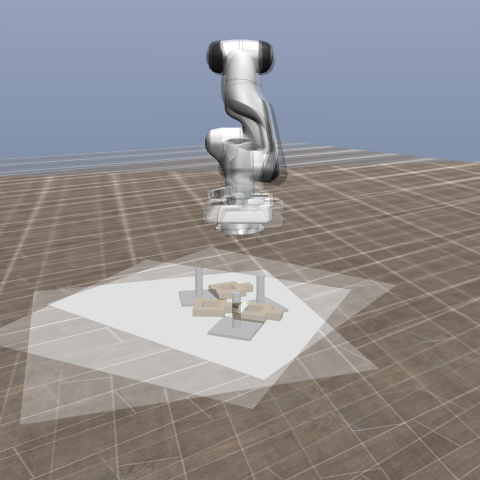}
      \caption*{Assembly Square}
    \end{subfigure} &
    \begin{subfigure}{0.15\textwidth}
      \includegraphics[width=\linewidth]{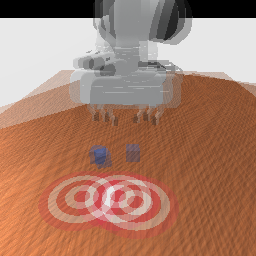}
      \caption*{Push}
    \end{subfigure} &
    \begin{subfigure}{0.15\textwidth}
      \includegraphics[width=\linewidth]{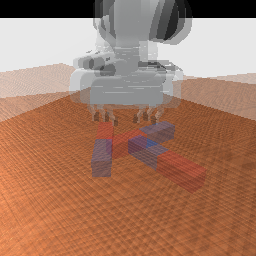}
      \caption*{Lift Upright}
    \end{subfigure} &
    \begin{subfigure}{0.15\textwidth}
      \includegraphics[width=\linewidth]{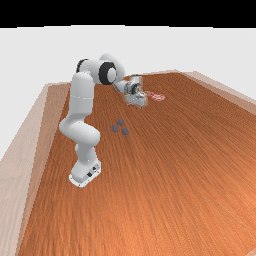}
      \caption*{Roll Ball}
    \end{subfigure}
  \end{tabular}
  \caption{Six custom tasks. The top row is the \textit{fixed} setups and the bottom row is the \textit{randomized} setups. The left three are in RoboSuite, and the right three are in ManiSkill. Each sub-figure overlays three images with different initialization seeds to illustrate variations in the environments.}
  \label{fig:tasks}
\end{figure*}
Another line of implicit methods focuses on learning view-robust representations through self-supervision. These approaches use cross-view correspondence~\cite{R3M2022, mvmwm}, contrastive learning~\cite{class}, or test-time adaptation~\cite{movie} to produce features that are invariant to viewpoint shifts without explicit geometric information.

In contrast to these implicit approaches, our work investigates the benefits of explicitly conditioning policies on camera geometry. By providing camera intrinsics and extrinsics directly to the model, we decouple the difficult task of pose inference from the primary goal of learning a manipulation policy, aiming for a more direct and data-efficient path to view invariance.

\subsection{Encoding Camera Geometries}
When camera geometry is known, various methods exist to encode it as an explicit input to a model. The dominant approach in multi-view transformers is to use ray-maps, which are pixel-aligned embeddings that provide geometric information at the token level. These ray-maps typically concatenate the 6D ray origin and direction for each pixel, derived from camera intrinsics and extrinsics~\cite{gao2024cat3d, mildenhall2021nerf}. A common variant uses 6D Plücker coordinates, which represent directed lines in space and offer the advantage of being invariant to the choice of origin along the ray~\cite{zhang2024cameras}.

To the best of our knowledge, our work is the first to introduce such camera embedding directly into RGB-based policies for robot control.





\section{Approach}

\subsection{\plucker Embedding as Camera Conditioning}

To provide the robot policy model with detailed information about camera geometries that is compatible with the pixel input space, we follow previous works from the 3D-vision community that condition the model with only input-level inductive biases for 3D reconstruction~\cite{yifan2022input, guizilini2022depth, vggt}. We use a per-pixel \emph{ray-map} \todotj{should we do emph to all ray-map in the paper?}, where each entry is a six-dimensional \plucker coordinate \mbox{$r = (d, m) \in \mathbb{R}^6$} representing the ray that projects to that pixel~\cite{Lanman2006, Sitzmann2021, vggt}.
A \plucker ray consists of a unit direction vector \( d \in \mathbb{R}^3 \) and a moment vector \( m = p \times d \), where \( p \) is any point on the ray (e.g., the camera center). 
This representation is homogeneous and satisfies the bilinear constraint \( d^\top m = 0 \), capturing oriented 3D lines in a uniform, continuous form. 
Given a camera with intrinsic matrix \( K \) and extrinsics (i.e., pose) \( (R, t) \), we compute the world-frame ray direction associated with pixel $(u,v)$ as
\begin{equation}
    d_w = \frac{R^\top K^{-1} \tilde{u}}{\|R^\top K^{-1} \tilde{u}\|}, \quad \text{where } \tilde{u} = [u, v, 1]^\top,
\end{equation}
and obtain the camera center as \( C = -R^\top t \). The full \plucker ray is then \( r = (d_w, C \times d_w) \). This defines a fixed mapping from each pixel to a ray in 3D space, independent of the scene content.

A challenge when integrating geometric conditioning is compatibility with policies that use pretrained vision encoders.
We propose two ways to encode the \( H \times W \times 6 \) ray-map (Fig.~\ref{fig:arch}), depending on whether the policy includes a pretrained visual encoder. For policies that do not include a pretrained encoder (e.g., Diffusion Policy~\cite{DiffusionPolicy2023}), we concatenate the \plucker map channel-wise with the image (Fig.~\ref{fig:arch-without-pretrained}). 
We modify the first layer of the image encoder such that it takes a $9$-channel rather than $3$-channel input and leave the rest of the policy network unchanged.
For policies with pretrained encoders, we employ a small convolutional network to encode the \plucker map to the same dimension as the image latent (Fig.~\ref{fig:arch-with-pretrained}). We then concatenate the output with the image latent channel-wise.

\subsection{Benchmarks and Task Setup}

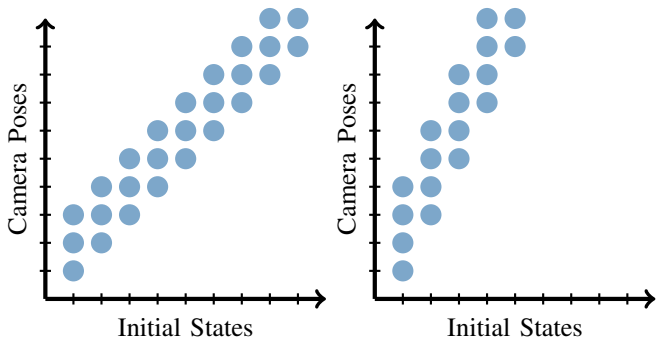
\begin{figure}
    \begin{subfigure}{0.48\columnwidth}
    \pgfmathsetmacro{\xMax}{10} 
    \pgfmathsetmacro{\yMax}{10}
    \pgfmathsetmacro{\xMaxMinusOne}{\xMax-1}
    \pgfmathsetmacro{\yMaxMinusOne}{\yMax-1}

    \pgfmathsetmacro{\xFirst}{1}
    \pgfmathsetmacro{\yFirst}{1}  
    \pgfmathsetmacro{\band}{3}    
    \pgfmathsetmacro{\yshift}{1}   
    \centering
    \begin{tikzpicture}[%
        x=\dimexpr0.9\columnwidth/\xMax\relax,
        y=\dimexpr0.9\columnwidth/\yMax\relax]
    
        \draw[->,ultra thick] (0,0)--(\xMax,0) node[right]{};
        \draw[->,ultra thick] (0,0)--(0,\yMax) node[above]{};
        %
        \node at ({\xMax/2},-1.0) {Initial States};
        \node[rotate=90] at (-1.0,{\yMax/2}) {Camera Poses};
        %
        %
        \foreach \x in {1,...,\xMaxMinusOne}
          \draw[thick] (\x,0.2) -- (\x,-0.2) node[below] {};
    
        \foreach \y in {1,...,\yMaxMinusOne}
          \draw[thick] (0.2,\y) -- (-0.2,\y) node[left] {};

        \pgfmathtruncatemacro{\ymin}{\yFirst}
        \foreach \x in {\xFirst,...,\xMax} {
            \pgfmathtruncatemacro{\ymax}{min(\ymin + \band - 1, \yMax)}
            \ifnum\ymin<\yMax
              \foreach \y in {\ymin,...,\ymax} {
                \fill[steelblue, opacity=0.7] (\x,\y) circle[radius=4pt];
              }
            \fi
            \pgfmathtruncatemacro{\nextymin}{min(\ymin + \yshift, \yMax)}
            \xdef\ymin{\nextymin}
          }
    \end{tikzpicture}
    \end{subfigure} \hfil
    \begin{subfigure}{0.48\columnwidth}
    \pgfmathsetmacro{\xMax}{10} 
    \pgfmathsetmacro{\yMax}{10}
    \pgfmathsetmacro{\xMaxMinusOne}{\xMax-1}
    \pgfmathsetmacro{\yMaxMinusOne}{\yMax-1}

    \pgfmathsetmacro{\xFirst}{1}
    \pgfmathsetmacro{\yFirst}{1}  
    \pgfmathsetmacro{\band}{4}    
    \pgfmathsetmacro{\yshift}{2}   
    \centering
    \begin{tikzpicture}[%
        x=\dimexpr0.9\columnwidth/\xMax\relax,
        y=\dimexpr0.9\columnwidth/\yMax\relax]
    
        \draw[->,ultra thick] (0,0)--(\xMax,0) node[right]{};
        \draw[->,ultra thick] (0,0)--(0,\yMax) node[above]{};
        %
        \node at ({\xMax/2},-1.0) {Initial States};
        \node[rotate=90] at (-1.0,{\yMax/2}) {Camera Poses};
        %
        %
        \foreach \x in {1,...,\xMaxMinusOne}
          \draw[thick] (\x,0.2) -- (\x,-0.2) node[below] {};
    
        \foreach \y in {1,...,\yMaxMinusOne}
          \draw[thick] (0.2,\y) -- (-0.2,\y) node[left] {};

        \pgfmathtruncatemacro{\ymin}{\yFirst}
        \foreach \x in {\xFirst,...,\xMax} {
            \pgfmathtruncatemacro{\ymax}{min(\ymin + \band - 1, \yMax)}
            \ifnum\ymin<\yMax
              \foreach \y in {\ymin,...,\ymax} {
                \fill[steelblue, opacity=0.7] (\x,\y) circle[radius=4pt];
              }
            \fi
            \pgfmathtruncatemacro{\nextymin}{min(\ymin + \yshift, \yMax)}
            \xdef\ymin{\nextymin}
          }
    \end{tikzpicture}
    \end{subfigure}
  \caption{Visualization of changes of two factors in data collection: camera pose and initial state of environment. On the left, $n = 3$ and $m = 1$. On the right, $n = 4$ and $m = 2$. \todointj{I think the blue looks better than this purple.}}
  \label{fig:2factor}
\end{figure}
We create three new variants of tasks in RoboSuite---Lift, Pick Place Can, Assembly Square---and three new variants of tasks in ManiSkill---Push, Lift Upright, Roll Ball---where we diversify the camera poses. We introduce two versions of each task as visualized in Figure~\ref{fig:tasks}. 
In the \textit{fixed} setting, the robot is fixed relative to the table and floor. In the \textit{randomized} setting, we randomize (i) the robot's position and orientation on the table; and (ii) the table position and orientation relative to the floor. 
The \textit{randomized} version emulates practical settings in which policies cannot exploit visual cues from the table or floor to (implicitly) estimate the pose of the camera relative to the robot. 

For our experiments, we intentionally exclude the wrist camera, as manipulation tasks often require information from third-person views that the wrist camera alone cannot provide. Doing so allows us to 
isolate the effects of camera conditioning on these third-person views and, in turn, policy performance.\todomw{I revisited the discussion of why we don't include wrist cameras}


Notice that when randomizing the camera poses during data collection, there are two factors being randomized: the camera pose and the initial state of the environment. For the model to efficiently generalize compositionally, we follow \citet{RobotDataComp2024} and collect our data in a ``stair''-shaped manner (Fig.~\ref{fig:2factor}): each demonstration episode is recorded with $n$ cameras, episodes $i$ and $i+1$ share $n-m$ camera poses, and episode $i+1$ involves $m$ new camera poses. For Can and Square, 
due to the difficulty of the task, we use $n=4$ and $m=2$. For the real-robot experiments and the rest of the simulated experiments, we use $n=3$ and $m=1$.
\begin{figure}[!t]
  \centering
  \setlength{\tabcolsep}{3pt}
  \begin{tabular}{cc}
    \begin{subfigure}{0.48\linewidth}
      \includegraphics[width=\linewidth]{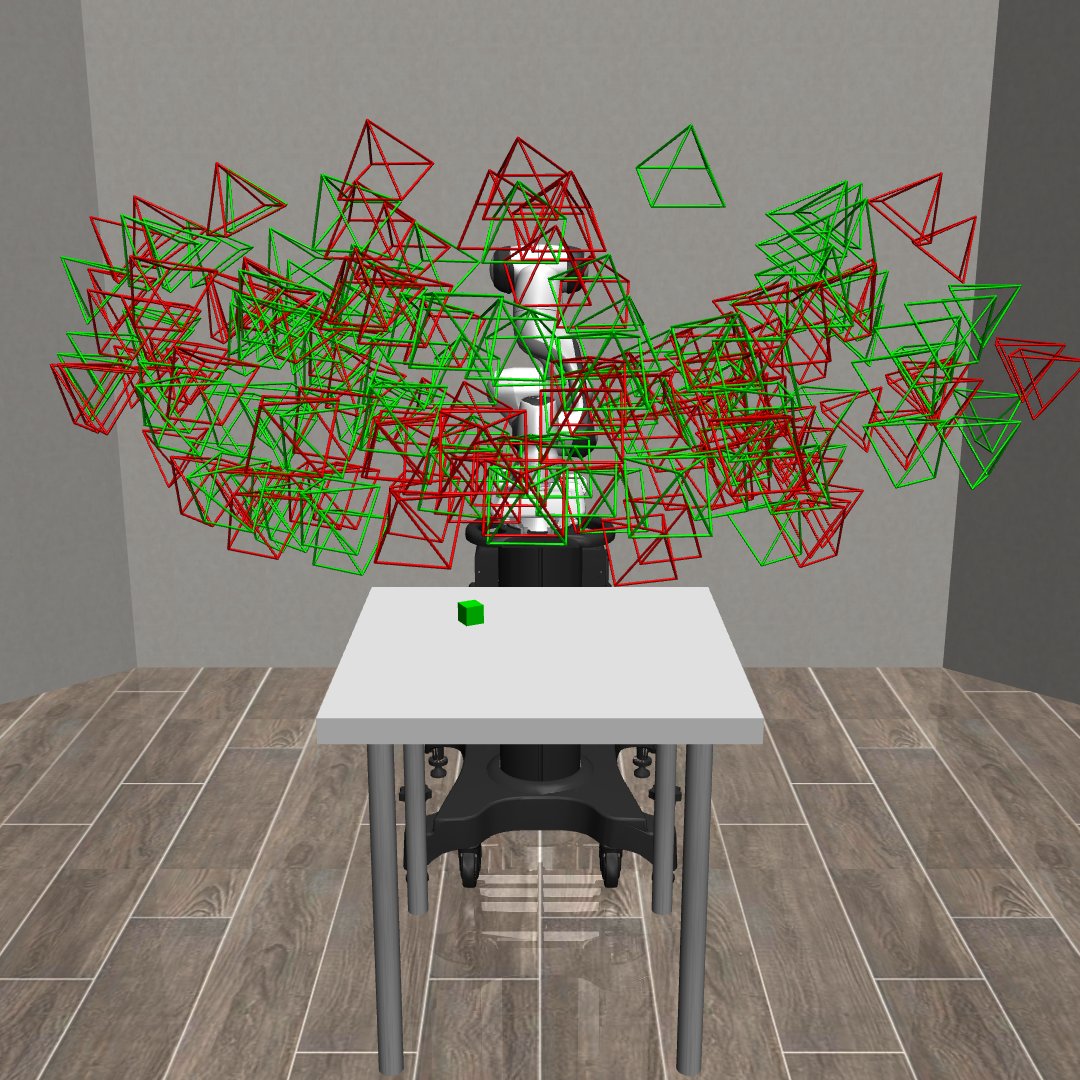}
      \caption*{Front View}
    \end{subfigure} &
    \begin{subfigure}{0.48\linewidth}
      \includegraphics[width=\linewidth]{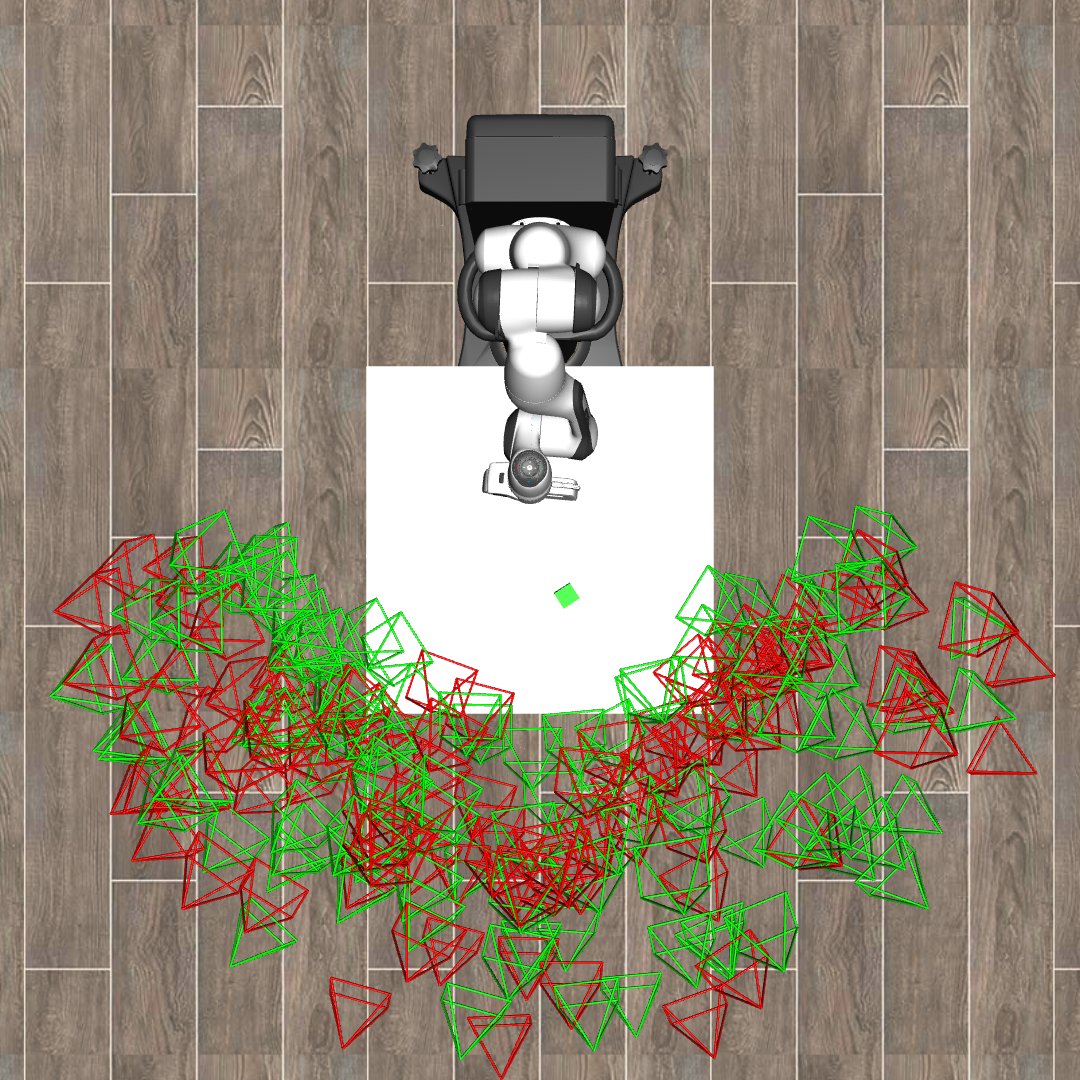}
      \caption*{Top View}
    \end{subfigure}
  \end{tabular}
  \caption{Visualization of camera poses. The training camera poses are in green and the test camera poses are in red.}
  \label{fig:sim_poses}
\end{figure}

The camera poses are visualized in Figure~\ref{fig:sim_poses}. For all tasks, we randomized the azimuth in a 180 degree range and elevation from 30 degrees to 60 degrees. We also randomized the 3D point that the cameras point to in a small 3D box.

\subsection{Evaluation Protocol}

We follow the evaluation protocol of \citet{DiffusionPolicy2023}: for each policy we train with three torch random seeds $(0, 1, 2)$, take the last ten checkpoints from each run, and evaluate every checkpoint on $50$ distinct initial position–camera pose pairs, yielding $3 \times 10 \times 50 = 1500$ rollouts per setting. 
For SmolVLA, due to compute constraints, we only run seed $0$, which yields $500$ rollouts per setting.
To ensure fairness, we fix the environment random seeds so that the initial conditions and stochastic environment noise are identical across checkpoints, torch seeds, and policies within each task setting. 
All the simulated experiments, including ablation studies, follow this evaluation protocol.

\section{Results} \label{sec:experiments}

We evaluate the robustness of behavior cloning policies to viewpoint variations and the benefits of pose conditioning through both simulation and real-world experiments. Simulation results are averaged across 1500 rollouts for ACT and Diffusion Policy and 500 rollouts for SmolVLA.

\subsection{Simulated Experiments}

\subsubsection{Camera Conditioning Increases Success Rate Across Models}
%
%
\begin{table*}[!t]
  \centering
  \caption{Success rates (\%) on simulated tasks with and without camera pose conditioning.}
  \label{tab:sim_success}
  {%
  \setlength{\tabcolsep}{8pt}
  \renewcommand{\arraystretch}{1.15}
      {
      \begin{tabularx}{\linewidth}{Xllllll}
        \toprule
        \textbf{Method} & \textbf{Lift} & \textbf{Pick Place Can} & \textbf{Assembly Square} & \textbf{Push} & \textbf{Lift Upright} & \textbf{Roll Ball} \\
        \midrule
        \textbf{ACT} & 33.6 & 26.7 & 10.8 & 29.9 & 22.9 & 28.7 \\
        \textbf{ACT w. conditioning} & 60.6 \g{(+27.0)} & 30.9 \g{(+4.2)} & 18.7 \g{(+7.9)} & 37.5 \g{(+7.6)} & 34.6 \g{(+11.7)} & 29.7 \g{(+1.0)} \\
        \midrule
        \textbf{DP} & 29.1 & 23.1 & \hphantom{0}2.0 & 20.0 & \hphantom{0}9.5 & 19.9 \\
        \textbf{DP w. conditioning} & 51.1 \g{(+22.0)} & 39.3 \g{(+16.2)} & \hphantom{0}2.4 \g{(+0.4)} & 30.3 \g{(+10.3)} & 20.7 \g{(+11.2)} & 23.7 \g{(+3.8)} \\
        \midrule
        \textbf{SmolVLA} & 19.6 & 56.0 & 22.0 & 39.0 & 23.6 & 27.6 \\
        \textbf{SmolVLA w. Conditioning} & 54.4 \g{(+34.8)} & 70.0 \g{(+14.0)} & 26.4 \g{(+4.4)} & 43.8 \g{(+4.8)} & 33.4 \g{(+9.8)} & 30.4 \g{(+2.8)} \\
        \bottomrule
      \end{tabularx}%
      }%
  }
\end{table*}
We train ACT~\cite{ACT2023}, Diffusion Policy~\cite{DiffusionPolicy2023}, and SmolVLA~\cite{SmolVLA2025} on the six tasks in simulation with and without camera conditioning. See Section~\ref{sec:simulation-training-details} for details. 
We see in Table~\ref{tab:sim_success} that adding camera pose conditioning increases success rates for all models in all tasks.

\begin{figure}[!t]
  \centering
  \includegraphics[width=\linewidth]{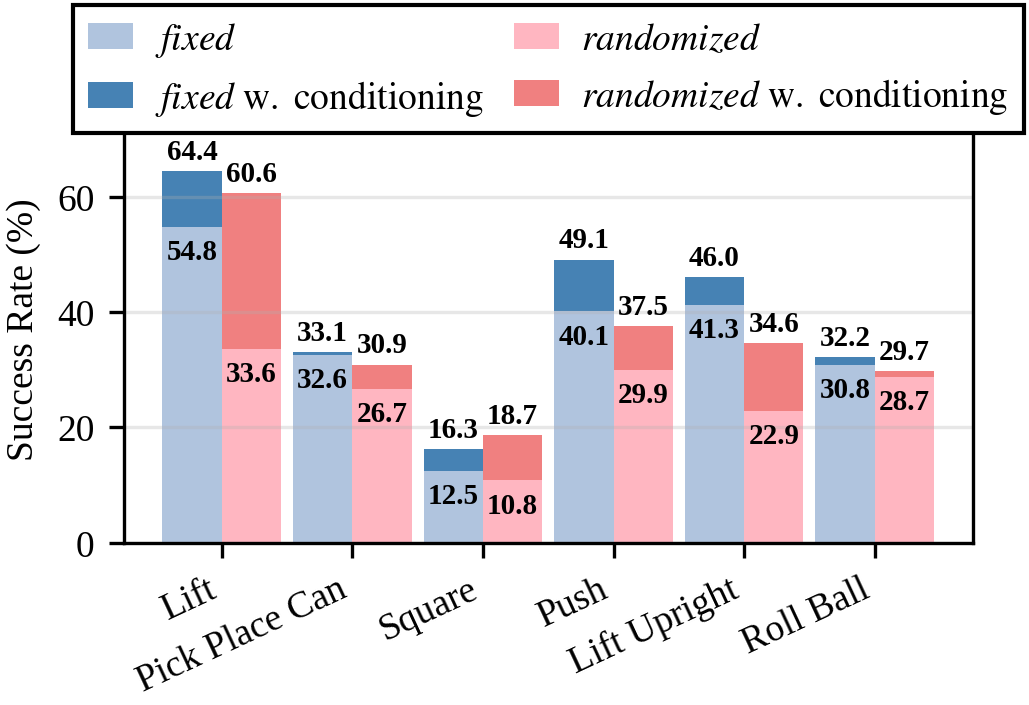}
  \caption{Success rates of policies for \textit{fixed} and \textit{randomized} settings, with and without \plucker conditioning.} 
  \label{fig:fixed_vs_rand}
\end{figure}

\subsubsection{Fixed vs.\ Randomized Background}
Figure~\ref{fig:fixed_vs_rand} evaluates the benefits of conditioning an ACT policy on camera extrinsics for the fixed and randomized variations of each task. Across all six tasks, the baseline ACT policy performs worse in the randomized setting, where it cannot exploit the environment structure to infer relative camera pose. It is in this setting that we see the greatest benefit of conditioning on camera pose; however, the results show that it is advantageous in the fixed setting as well.



\subsubsection{Random Cropping} 
\begin{figure}[h]
  \centering
  \includegraphics[width=\linewidth]{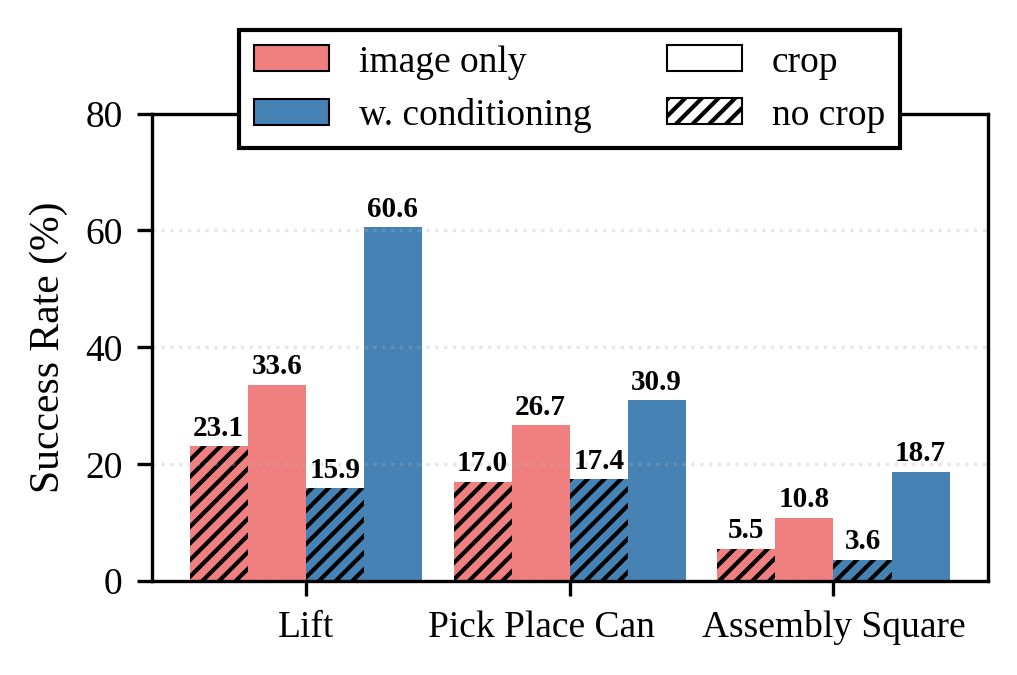}
  \caption{Ablation of random cropping's effect on task success.}
  \label{fig:cropping}
\end{figure}

Random cropping is applied to both the \plucker maps and the images, ensuring that the one-to-one correspondence between image pixels and \plucker ``pixels'' is preserved. 
Figure~\ref{fig:cropping} shows the success rates of the three RoboSuite tasks with and without random cropping.
Cropping consistently improves performance across all tasks. 
One interpretation of the results is that cropping effectively adds virtual cameras with different parameters.

\subsubsection{Action Space} 
\begin{figure}[h]
  \centering
  \includegraphics[width=\linewidth]{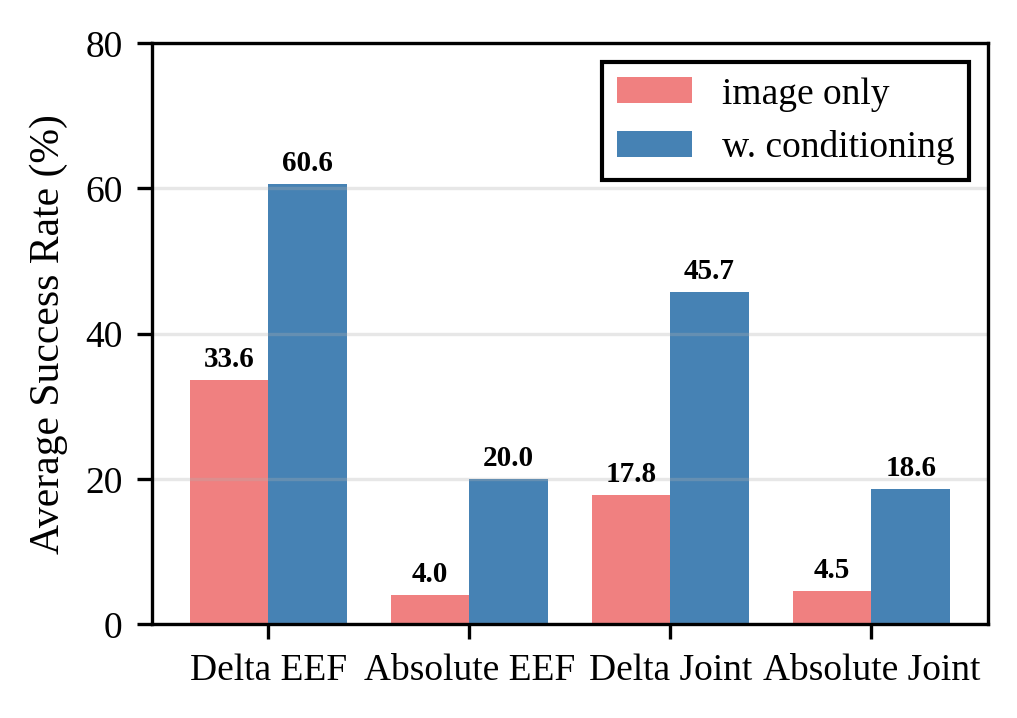}
  \caption{Success rates for different action spaces.} 
  \label{fig:action_space}
\end{figure}

We ablate across four action spaces: Absolute End-Effector Pose, Delta End-Effector Pose, Absolute Joint Position, and Delta Joint Position. 
To ensure fairness, we use the same set of demonstrations, converting them into each action space, and verify by replay that all achieve success rates above 99 percent.
Delta actions share the same underlying absolute controllers: Delta End-Effector Pose is derived by differencing adjacent absolute poses and accumulating at test time, while Delta Joint Position is derived analogously from Absolute Joint Position. 
Thus, any differences between delta and absolute variants reflect the policies’ generalization, not the controllers themselves. 
We see in Figure~\ref{fig:action_space} that the policy performs best when actions take the form of Delta End-Effector Pose and that the performance is noticeably worse for Absolute Joint Position and Absolute End-Effector Pose.
Nonetheless, the results reveal that camera conditioning increases performance for all four action spaces.

\begin{figure}[h]
  \centering
  \includegraphics[width=\linewidth]{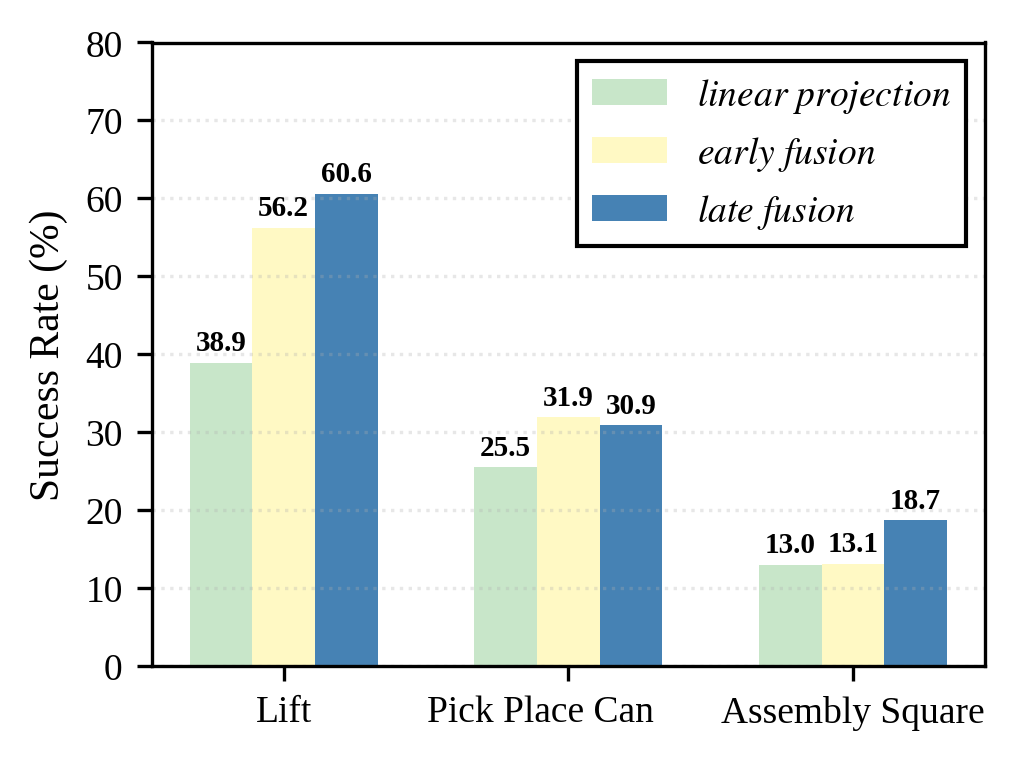}
  \caption{Ablation of different camera pose encoding methods.}
  \label{fig:encoding}
\end{figure} 

\subsubsection{Different Ways to Encode Camera Poses} 
We ablate different approaches to encoding camera poses, as shown in Figure~\ref{fig:encoding}. 
In the \textit{linear projection} method, the extrinsic matrix is linearly projected to a token. 
In the \textit{early fusion} method, the \plucker map is concatenated with the image pixel-wise before the pretrained encoder. 
In the \textit{late fusion} method, the \plucker and image latents are concatenated channel-wise after separate encoders, which is the approach we adopt for ACT and SmolVLA in the main experiments. 
We find that the \textit{linear projection} method provides no observable benefit. 
In most cases, \textit{early fusion} is less effective than \textit{late fusion}. 
This may be because concatenating the \plucker map with the image creates inputs that are out of distribution for the pretrained ResNet at the start of training, whereas encoders trained from scratch do not face this issue.

\subsubsection{The Effect of Pretrained Vision Encoders}
\begin{figure}[h]
  \centering
  \includegraphics[width=\linewidth]{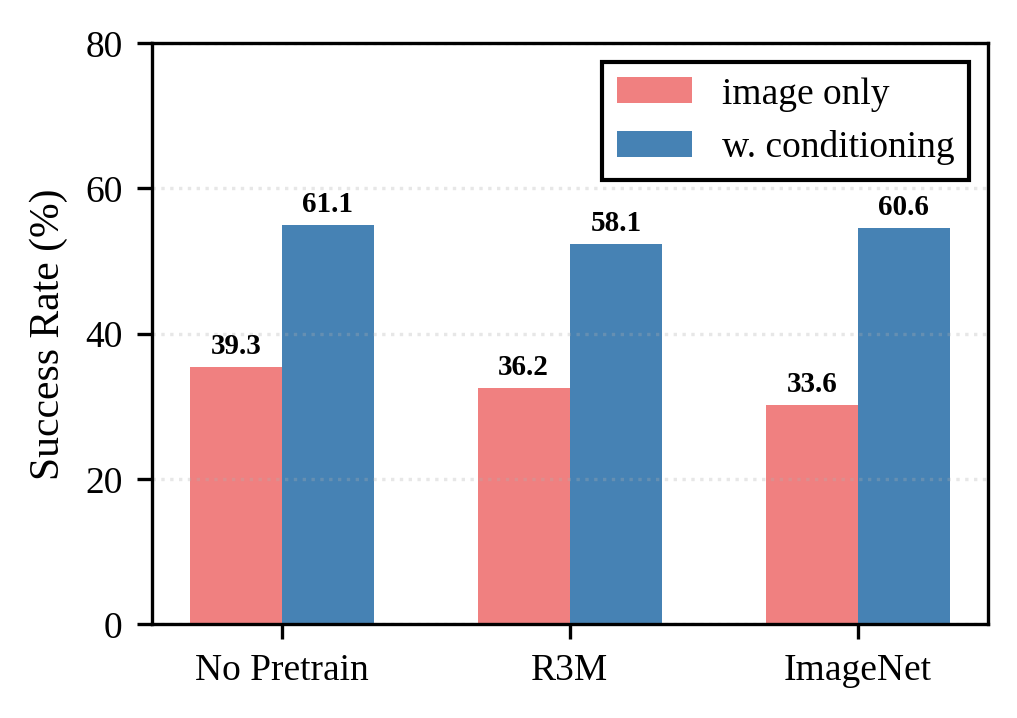}
  \caption{Ablation on pretraining of image encoders.}
  \label{fig:pretrain}
\end{figure}

We find that pretraining the image encoder (Fig.~\ref{fig:pretrain}) has little effect on the success rate. 
We trained ACT with ResNet encoders under three settings: no pretraining, R3M weights, and pretraining on ImageNet. 
Based on these results, we follow the original papers in our experiments: ImageNet-pretrained ResNet for ACT, randomly initialized ResNet for Diffusion Policy, and SigLIP \cite{siglip} for SmolVLA.

\begin{figure}[htbp]
  \centering
  \includegraphics[width=\linewidth]{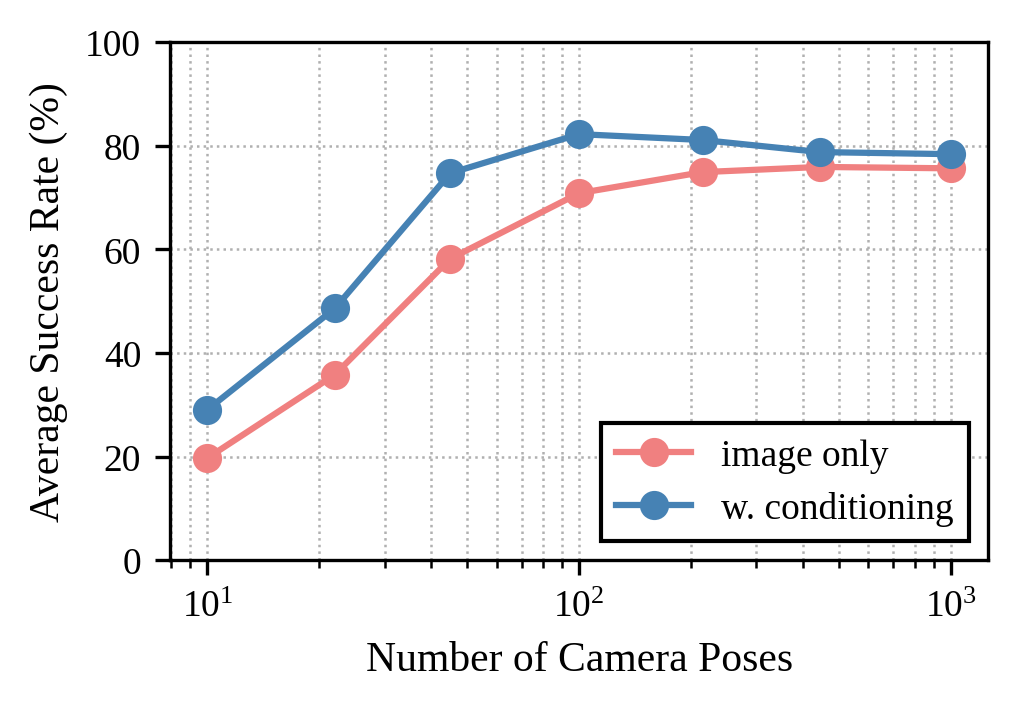}
  \caption{Experiment on scaling to more camera poses with the Pick task in RoboSuite.}
  \label{fig:scaling}
\end{figure}

\subsubsection{Scaling}
We evaluate the benefits of conditioning on camera pose when scaling the training data to more views. For these experiments, all demonstrations share the same set of $n$ cameras ($m=0$). The results are shown in Figure~\ref{fig:scaling}.
The $x$-axis shows the number of cameras on a logarithmic scale.
We observe that, to achieve the same performance, training without camera pose conditioning requires several times more cameras compared to training with pose conditioning.
Moreover, as the number of cameras increases further, conditioning on camera pose still provides a small but consistent improvement.

\subsubsection{Training Details} \label{sec:simulation-training-details}
ACT and SmolVLA are trained for 30k epochs and evaluated every 1k epochs. 
Diffusion Policy is trained for 80k epochs and evaluated every 2k epochs.
The number of epochs is selected so that the last 10 evaluation starts\todomw{Grammar: ``the last 10 evaluation starts''} well after the policies' success rates have converged.
ACT and Diffusion Policy use a constant learning rate schedule.
SmolVLA starts from the pretrained weights from LeRobot~\cite{lerobot-smolvla} and we train it with a linear warmup and a cosine decay of learning rate, as we find it is difficult to optimize with a constant learning rate schedule.

\subsection{Real-World Experiments}

\begin{figure}[!t]
  \centering
  \setlength{\tabcolsep}{3pt} 
  \begin{tabular}{ccc}
    \begin{subfigure}{0.31\linewidth}
      \includegraphics[width=\linewidth]{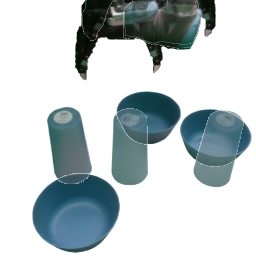}
      \caption*{Pick Place}
    \end{subfigure} &
    \begin{subfigure}{0.31\linewidth}
      \includegraphics[width=\linewidth]{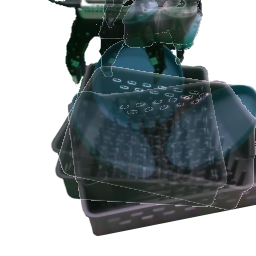}
      \caption*{Plate Insertion}
    \end{subfigure} &
    \begin{subfigure}{0.31\linewidth}
      \includegraphics[width=\linewidth]{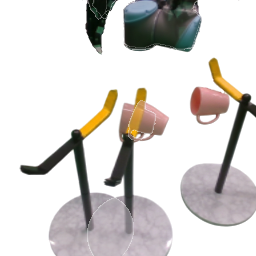}
      \caption*{Hang Mug}
    \end{subfigure}
  \end{tabular}
  \caption{Real-robot tasks. Each sub-figure overlays three images with different initial states, to illustrate variations in the tasks.}
  \label{fig:real_tasks}
\end{figure}

\begin{figure}[h]
  \centering
  \includegraphics[width=\linewidth]{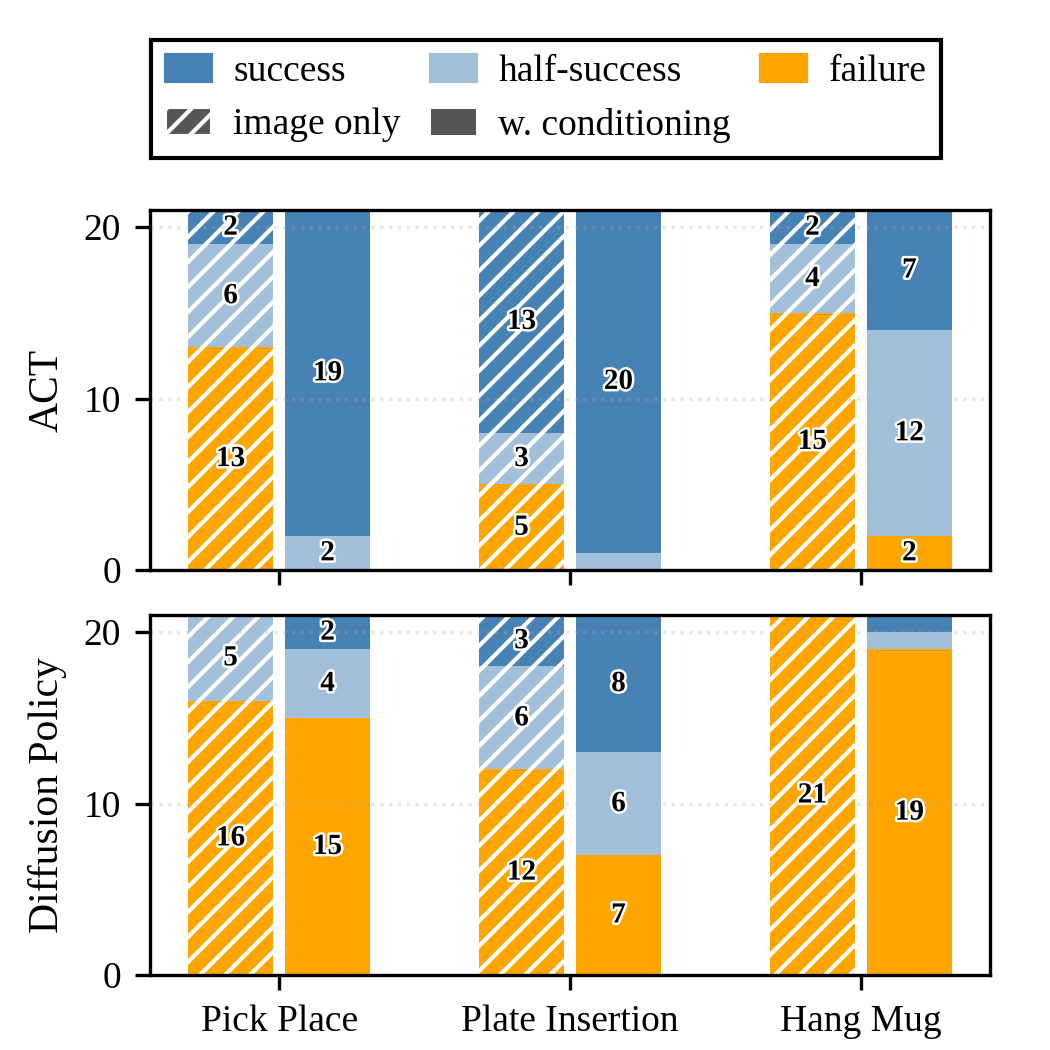}
  \caption{The performance of (top) ACT and (bottom) Diffusion Policy with and without camera pose conditioning for the three real-world experiments with 21 rollouts per setting.\protect\footnotemark}
  \label{fig:real_plot}
\vspace{-3mm}
\end{figure}

We conduct real-robot experiments using a UR5 robot arm (Fig.~\ref{fig:real_robot}) equipped with a Robotiq three-finger gripper on three tasks (Fig.~\ref{fig:real_tasks}): Pick Place, Plate Insertion, and Hang Mug.

For each task, we collect $200$ demonstrations using a VR-based teleoperation interface.
Three cameras are mounted on articulated camera arms, allowing us to vary their poses.
We follow the same randomization strategy as in the simulation experiments: each demonstration is recorded with three cameras ($n=3$), and after each recording, we change the pose of one camera ($m=1$). We estimate the pose of each camera using AprilTags~\cite{wang2016apriltag}. 
We train ACT for 30k epochs and Diffusion Policy for 60k epochs.

We aim to replicate the \textit{randomized} setting employed in the simulation benchmarks, which is consistent with practical policy deployments.
Because the UR5 is heavy and difficult to reposition, we approximate this setting by covering the background with green cloth to remove static visual cues.
\footnotetext{We observed that the performance of Diffusion Policy is significantly worse than that of ACT. We believe this is partly due to its stochastic nature in precision-demanding tasks. In particular, when multiple trajectories can complete a task, Diffusion Policy often oscillates between them, which can lead to failure.}

For evaluation, we test with seven random poses for each of the three cameras.
In each setting, we run one rollout using two of the three cameras, resulting in $3 \times 7 = 21$ rollouts per setting.
To ensure fairness, we mark the initial positions of objects and reset them consistently across all methods.\todomw{Changed from ``settings'' to ``methods''}

Due to the difficulty of the real-robot tasks, we introduce a criterion called half-success. 
This allows us to provide more nuanced information that is not captured by a simple binary measure of success or failure.
We define half-success separately for each task as follows:
\begin{itemize}
    \item \textbf{Pick Place:} Half-success means the robot attempts to lift the cup, but it slips when the grasp is slightly off. Success means lifting the cup and placing it in the bowl.
    \item \textbf{Plate Insertion:} Half-success means the robot successfully picks up the plate. Success requires inserting the plate into the slots of the dish rack.
    \item \textbf{Hang Mug:} Half-success occurs when the mug makes contact with the hanger. Success requires that the mug ends up hanging securely on the hanger.
\end{itemize}

We see from Figure~\ref{fig:real_plot} that conditioning on camera poses improves all policies for all tasks.

\section{Conclusion}

We studied view-invariant imitation learning by conditioning policies on camera extrinsics using Plücker ray-maps.
Our experiments show consistent improvements in viewpoint generalization across ACT, Diffusion Policy, and SmolVLA in both simulation and real robot settings.
We also highlighted how policies without extrinsics can exploit static background cues and fail when those cues change, while extrinsics conditioning restores performance.

Despite these encouraging results, our study has its limitations. Current pose estimation methods remain imperfect, particularly in scenarios with featureless surfaces, limited view overlap, or highly dynamic scenes. In such cases, pose estimation errors can compound with downstream control if no mitigation strategies are applied. Nonetheless, rapid progress in generalized pose estimation~\cite{huang2025vipe} suggests this sub-problem may be addressed before the broader challenge of end-to-end robot policy learning is solved. Moreover, our work mainly focuses on pose conditioning, while many other directions remain unexplored. For example, just as our method enables policies to generalize to novel camera poses, a natural extension would be to support generalization across cameras with different intrinsics.

Looking forward, we believe these contributions will provide both methodological insights and practical benchmarks for advancing viewpoint-robust imitation learning. In particular, our six new RoboSuite and ManiSkill benchmarks offer a principled way to evaluate robustness, while our findings on action space design and data augmentation highlight key considerations for policy training. Together, these contributions lay the foundation for developing policies that generalize more reliably across the diverse viewpoints encountered in real-world deployment.

{\small 
\balance
\bibliographystyle{IEEEtranN}
\bibliography{references}
}

\clearpage

\end{document}